\documentclass[a4paper, 10pt, conference]{ieeeconf}        

\IEEEoverridecommandlockouts                              
\usepackage[utf8]{inputenc}
\usepackage[english]{babel}

\usepackage{cite}
\usepackage{multirow}
\usepackage{graphicx} 
\usepackage[table,xcdraw]{xcolor}
\usepackage{epsfig} 

\usepackage{amsmath} 
\usepackage{amssymb}  
\usepackage{amsthm}

\theoremstyle{plain}

\usepackage{flafter} 

\usepackage{caption}
\usepackage{subcaption}

\usepackage{algorithm}
\usepackage[]{algpseudocode} 

\usepackage{program}

\usepackage{breqn}

\usepackage{siunitx} 
\usepackage{gensymb} 

\usepackage{tabularx}

\usepackage{hyperref}
\usepackage{cleveref}

\newcolumntype{L}[1]{>{\raggedright\arraybackslash}p{#1}}
\newcolumntype{C}[1]{>{\centering\arraybackslash}p{#1}}
\newcolumntype{R}[1]{>{\raggedleft\arraybackslash}p{#1}}

\usepackage[nolist,nohyperlinks]{acronym}

\graphicspath{{figures/}}

\title{\LARGE \bf A Multi-Segment, Soft Growing Robot with Selective Steering}%

\author{Alexander M. Kübler$^{1,2}$, Sebastián Urdaneta Rivera$^{1}$, Frances B. Raphael$^{1}$,  Julian Förster$^{2}$, \\ Roland Siegwart$^{2}$, and Allison M. Okamura$^{1}$ 
\thanks{The project was supported in part by the U.S. Department of Energy, National Nuclear Security Administration, Office of Defense Nuclear Nonproliferation Research and Development (DNN R\&D) under subcontract from Lawrence Berkeley National Laboratory; and the United States Federal Bureau of Investigation contract 15F06721C0002306.}
\thanks {$^{1}$CHARM Lab, Dept. of Mechanical Engineering, Stanford University, Stanford, CA 94305, USA. Email: \{akuebler, sebasur, fraphael, aokamura\}@stanford.edu}
\thanks{$^{2}$Autonomous Systems Lab, Dept. of Mechanical and Process Engineering, ETH Zürich, 8092 Zürich, Switzerland. Email: \{akuebler, fjulian, rsiegwart\}@ethz.ch}
}

\begin{document}

\maketitle
\thispagestyle{empty}
\pagestyle{empty}


\begin{abstract}
Everting, soft growing vine robots benefit from reduced friction with their environment, which allows them to navigate challenging terrain. Vine robots can use air pouches attached to their sides for lateral steering. However, when all pouches are serially connected, the whole robot can only perform one constant curvature in free space. It must contact the environment to navigate through obstacles along paths with multiple turns.
This work presents a multi-segment vine robot that can navigate complex paths without interacting with its environment. This is achieved by a new steering method that selectively actuates each single pouch at the tip, providing high degrees of freedom with few control inputs.
A small magnetic valve connects each pouch to a pressure supply line. A motorized tip mount uses an interlocking mechanism and motorized rollers on the outer material of the vine robot. As each valve passes through the tip mount, a permanent magnet inside the tip mount opens the valve so the corresponding pouch is connected to the pressure supply line at the same moment. 
Novel cylindrical pneumatic artificial muscles (cPAMs) are integrated into the vine robot and inflate to a cylindrical shape for improved bending characteristics compared to other state-of-the-art vine robots.
The motorized tip mount controls a continuous eversion speed and enables controlled retraction. A final prototype was able to repeatably grow into different shapes and hold these shapes. We predict the path using a model that assumes a piecewise constant curvature along the outside of the multi-segment vine robot. The proposed multi-segment steering method can be extended to other soft continuum robot designs.
\end{abstract}


\section{Introduction} \label{sec:introduction}

\begin{figure}[t]
    \centering
    \includegraphics[width=0.48
    \textwidth]{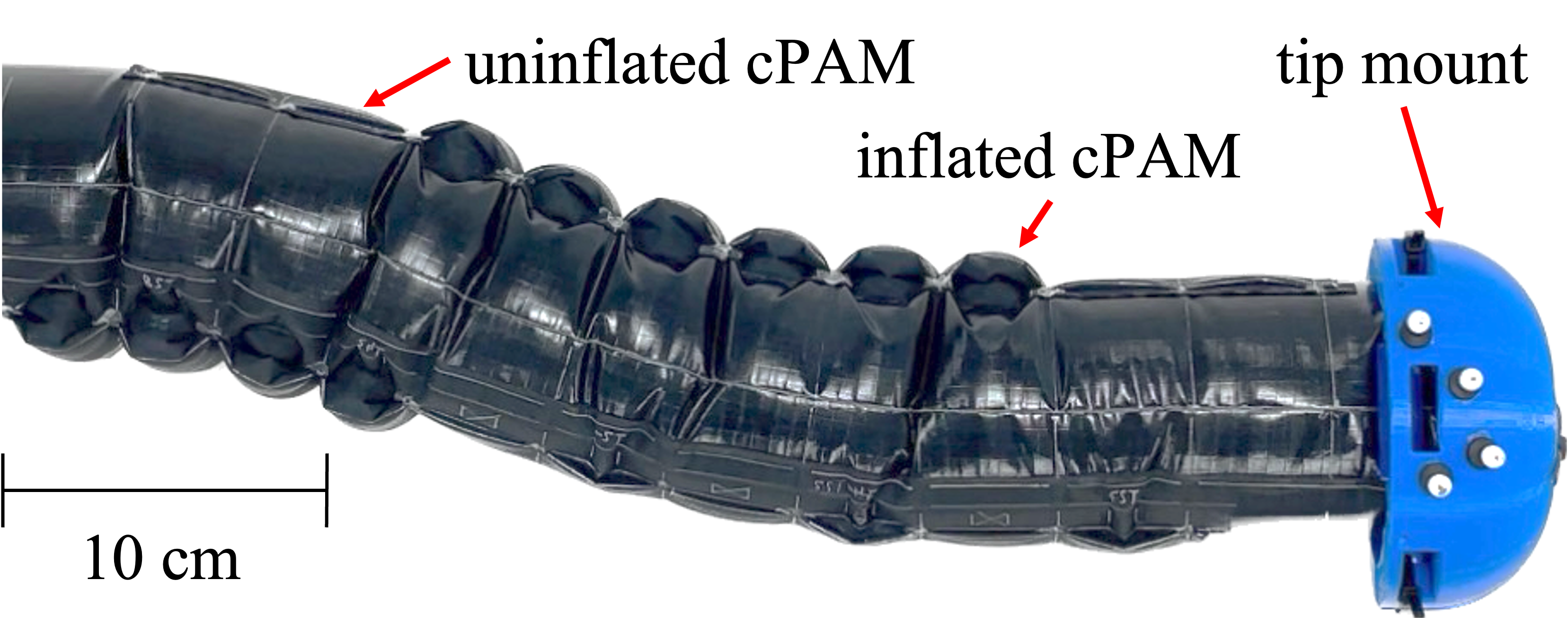}
    \caption{A multi-segment vine robot with selectively actuated cylindrical pneumatic artificial muscles (cPAMs) and a motorized tip mount with permanent magnets for cPAM valve control.}
    \label{fig:firstvine}
\end{figure}

\subsection{Motivation}

Soft continuum robots offer great opportunities due to their infinite degrees of freedom and embodied intelligence \cite{rus2015design}.
Pneumatic, soft growing robots, so-called vine robots, are a specific kind of soft continuum robot with the ability to steer and move through growth. They consist of an inverted soft fabric or plastic tube that grows via tip extension when pressurized. This process is referred to as eversion \cite{hawkes2017soft}. Vine robots experience minimal friction with their environment because the outer material does not slide with respect to the surface. This makes vine robots a great choice to maneuver through challenging and dangerous environments. 

Applications of vine robots include exploration \cite{coad2019vine}\cite{luong2019eversion}, search and rescue \cite{der2021roboa}, medicine \cite{hwee2021everting}, and haptics \cite{agharese2018hapwrap}. 

Steering of vine robots is often achieved by pressurizing serially connected pouches on the outside of the vine robot body. Two antagonistically placed series of pouches will achieve 2D bending, and three will achieve 3D bending\cite{coad2019vine}. Because all pouches are connected in series, the vine robot can only achieve one constant curvature in free space.
To navigate around obstacles and achieve complex paths, current vine robots need to push from their environment \cite{selvaggio2020obstacle}. This is undesirable in unstable and dangerous environments or in environments with sharp edges and holes. To surmount these limitations, we propose a multi-segment vine robot that can achieve complex shapes without relying on environmental contact (\Cref{fig:firstvine}). Our vine robot can move around obstacles and can follow a specific path in free space, e.g., for data acquisition or exploration. Compared to the series pouch design, our proposed solution better leverages soft continuum robots' infinite degrees of freedom and only requires a few additional parts in its body that do not prevent eversion.

\subsection{Related Work}

The process of robot growth is inspired by vines. It can be mimicked using inverted fabric or plastic tubes, which evert and grow when pressurized \cite{greer2019soft}. The eversion can be modeled using biological apical extension \cite{blumenschein2017modeling}, or ideal gas assumptions \cite{franco2022model}. 

Different methods have been developed to steer a vine robot. Common solutions steer the vine robot over its full length using tendons\cite{blumenschein2018helical}, pouch motors \cite{coad2019vine}, fPAMs \cite{naclerio2020simple}, or sPAMs \cite{greer2019soft} attached to the side of the robot. Other steering mechanisms only change the shape of the vine robot's frontmost segment \cite{der2021roboa} \cite{haggerty2021hybrid}. 

Previous work on multi-segment vine robots includes preformed designs \cite{blumenschein2017modeling}, shape-locking techniques \cite{wang2020dexterous}, latches unleashed using pressure \cite{hawkes2017soft}, or the stiffening of certain segments combined with tendon steering \cite{do2020dynamically}. These methods either do not allow for flexible deployment (preformed), need an additional locking or stiffening system besides actuation (shape-locking, stiffening), or are not reversible and difficult to manufacture (latches).

A main challenge in the design of vine robots is the attachment of a tip mount to facilitate the transportation of a payload (e.g., sensors), because the tip of the vine robot is constantly changing. Furthermore, vine robots are prone to buckle unpredictably during retraction, requiring a design that prevents buckling \cite{coad2020retraction}. Tip mounts without retraction capability include a cap on the front of the vine robot \cite{coad2019vine}, a camera on a wire through the inside of the vine robot \cite{greer2019soft}, or the combination of an internal and external tip mount connected by magnetic \cite{luong2019eversion}, or interlocking \cite{der2021roboa} rollers. Tip mounts that also retract the vine robot use motorized rollers on the inner material of the vine robot \cite{coad2020retraction}\cite{jeong2020tip} or linear soft actuators \cite{heap2021soft}.

\subsection{Contributions}
The contributions of this work are:
\begin{itemize}
    \item A high-degree-of-freedom steering method for soft continuum robots that uses magnetic valves to actuate air pouches selectively. In this work, the method was implemented on a vine robot that can grow into different shapes.
    \item A 3D-printed, low-cost, magnetic, and externally controlled pneumatic valve, which can be tuned for different working ranges.
    \item A cylindrical, folded pouch design integrated into the vine robot body, which allows for high bending angles, the cylindrical pneumatic artificial muscle (cPAM).
    \item A motorized tip mount with an interlocking mechanism and motorized rollers working on the outer material of the vine robot to enable a continuous eversion speed and prevent buckling during retraction.
    \item A kinematic model for multi-segment vine robots using a piecewise constant curvature assumption.
\end{itemize}

\section{Design}
\label{sec:design}

\subsection{Multi-Segment Working Principle}

\begin{figure}[t]
    \centering
    \includegraphics[width=0.48
    \textwidth]{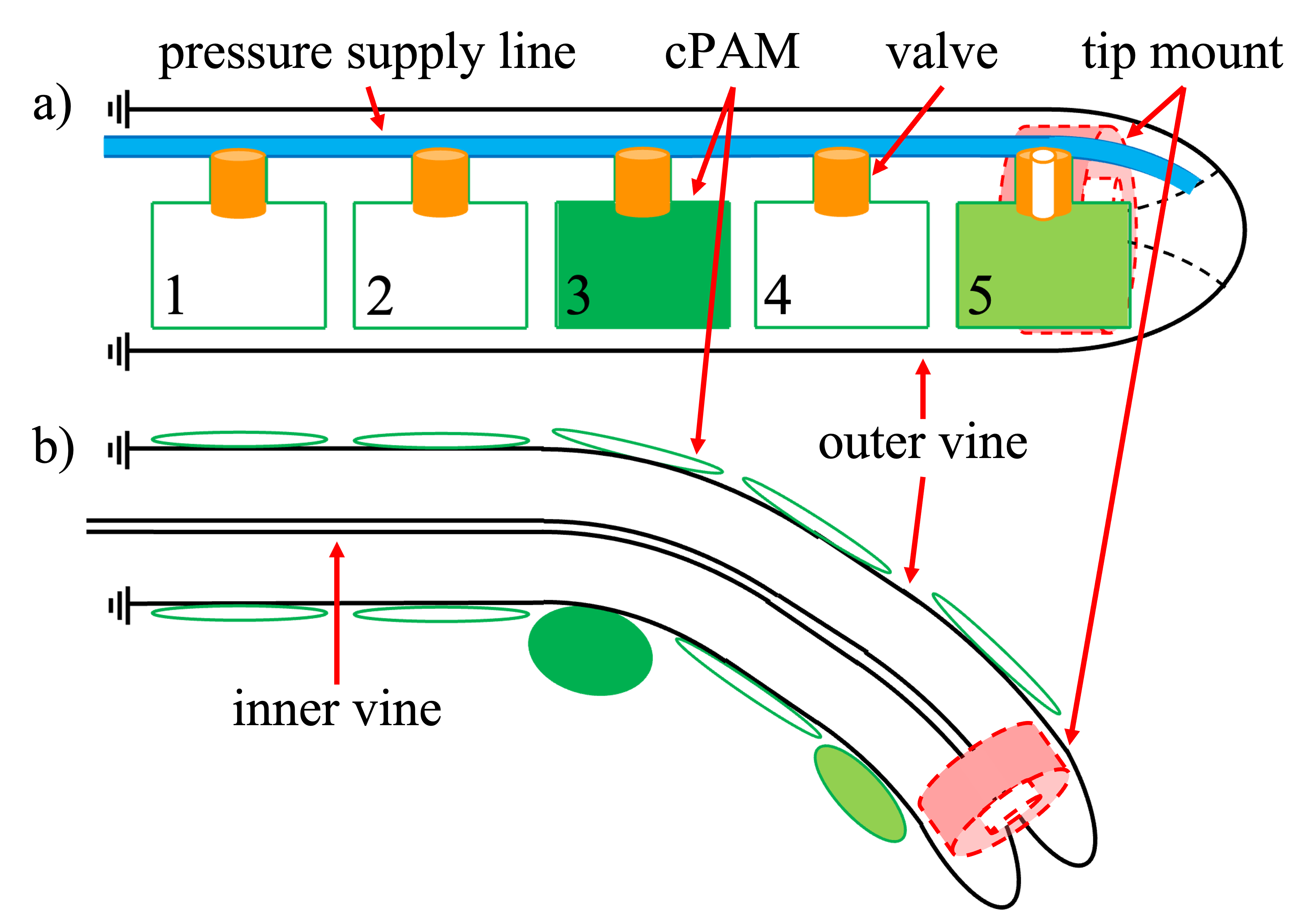}
    \caption{The main working principle from a) a side view and b) a top view. Green represents the pressure in the pouch (darker green means higher pressure). Individual valves only open near the tip mount. In the position shown, valve 5 is open and pouch 5 can be controlled using the pressure supply line. Valves 1-4 are closed and hold the previously applied pressure: No pressure was applied to pouches 1, 2, and 4, and the pressure applied to pouch 3 was higher than the one applied to pouch 5. This leads to a large bend to the right at pouch 3 and a smaller bend at pouch 5.}
    \label{fig:principle}
\end{figure}

\begin{figure}[t]
    \centering
    \includegraphics[width=0.3
    \textwidth]{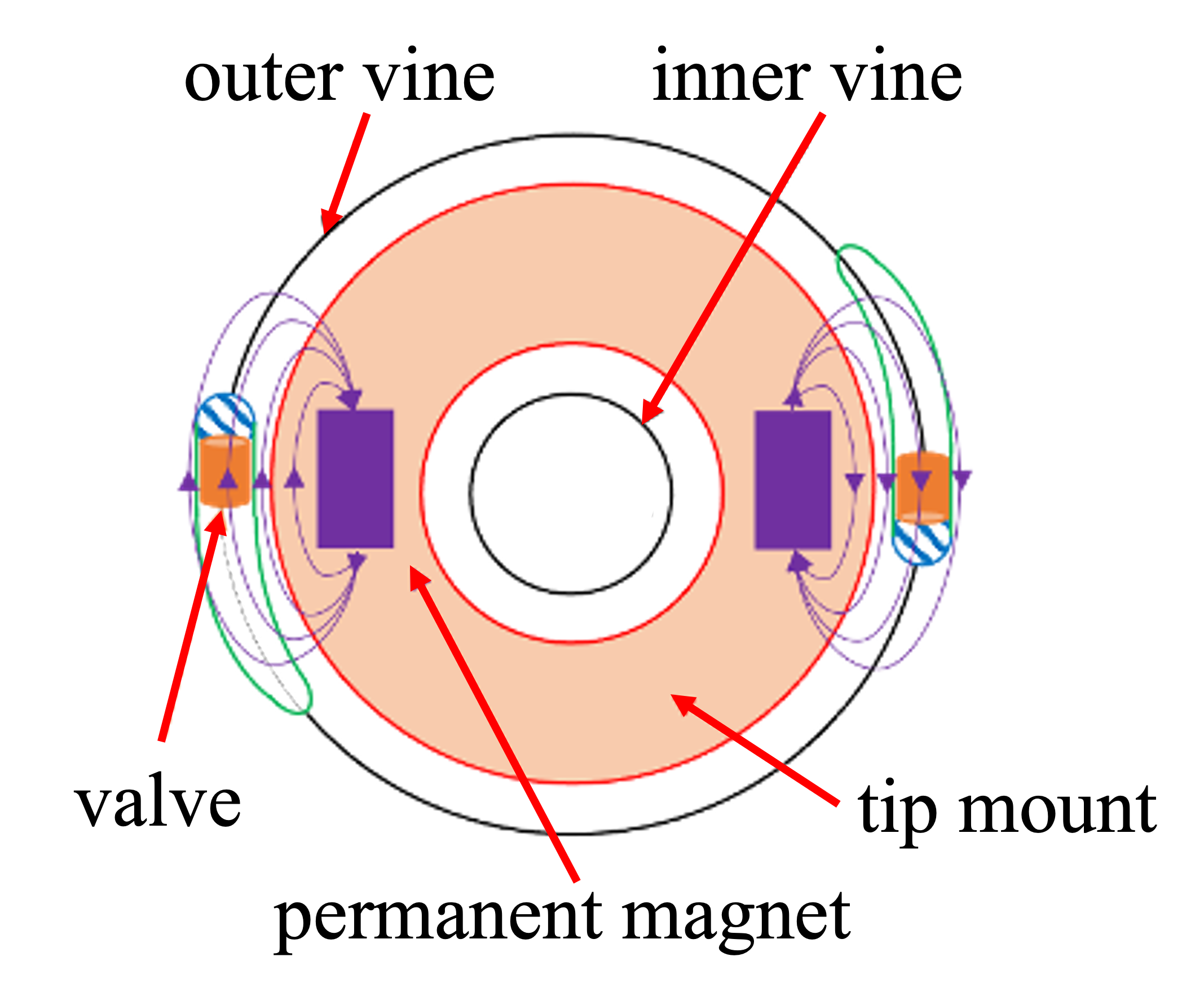}
    \caption{Schematic front view of the vine robot's tip. Permanent magnets inside the tip mount actuate the magnetic valves.}
    \label{fig:tip_principle}
\end{figure}

To allow a vine robot to grow into various shapes, every pouch (or a set of connected pouches) has to be actuated individually. \Cref{fig:principle} shows the principle. Instead of connecting all pouches in series, they are connected in parallel to a pressure supply line, with each pouch having its own magnetic valve. This valve is normally closed and only opens in the presence of an external magnetic field. Such a field can be generated at the tip mount, i.e., at the front of the vine robot, by placing a permanent magnet in the tip mount as shown in \Cref{fig:tip_principle}. 
With this principle, we can selectively actuate the frontmost pouch. Furthermore, when retracting the vine robot, the valves open again and the pouches are deflated by applying a vacuum to the pressure supply line. The pouches can only inflate after they are everted at the tip, in contrast to pouches connected in series. This enables eversion and growth at lower pressures because deflated pouches are easier to evert.

\subsection{Valve Design}

\begin{figure}[t]
    \centering
    \includegraphics[width=0.48
    \textwidth]{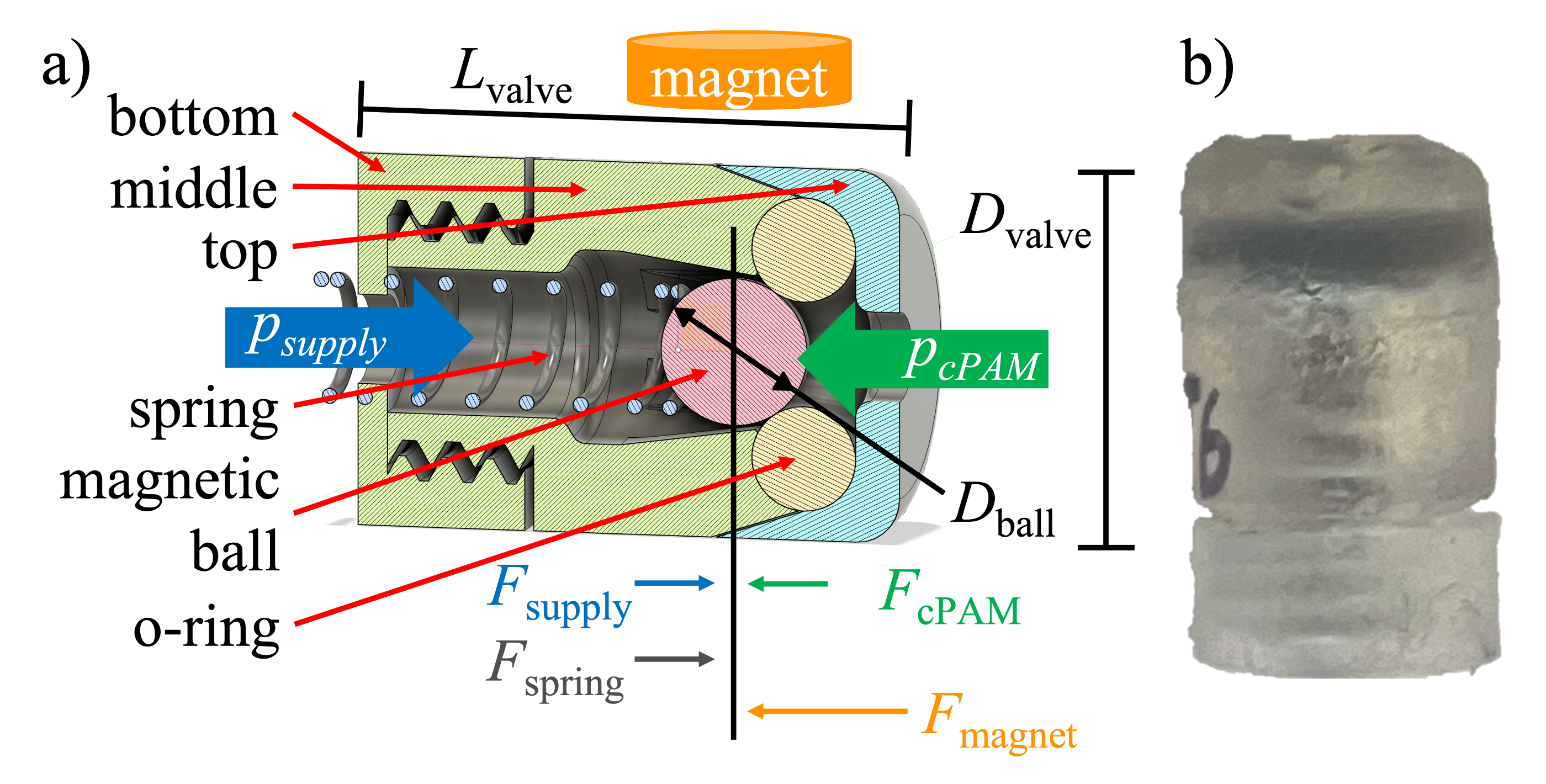}
    \caption{a) Inside view of the magnetic valve with the 3D-printed housing, the spring-loaded magnetic ball, and the o-ring. The forces acting on the magnetic ball result from the spring, the external permanent magnet, and the pressure in the pressure supply line and the cPAMs. The dimensions of the valve are $L_{\text{valve}} = 11$\,mm, $D_{\text{valve}} = 7$\,mm, and $D_{\text{ball}} = 2.5$\,mm. b) A manufactured valve.} 
    \label{fig:valve}
\end{figure}

Each magnetic valve consists of a 3D-printed housing, a spring-loaded (W.B. Jones) magnetic neodymium ball (Speks), and an o-ring (McMaster-Carr). \Cref{fig:valve} shows the inside of the valve and the acting forces. Pressure can act on the magnetic ball from both sides, from the pressure supply line~($F_{\text{supply}}$) and from the cPAM~($F_{\text{cPAM}}$). $F_{\text{supply}}$ is maximal if $p_{\text{supply}} = p_{\text{max}}$ and $F_{\text{cPAM}}$ is maximal if $p_{\text{cPAM}} = p_{\text{max}}$. These forces always act on all valves, not just the frontmost ones. The maximal pressure of the system was designed to be $p_{\text{max}} = 40$ kPa as this pressure allows for almost maximal bending. The valve must remain closed when a magnet is not present, independent of whether $F_{\text{supply}}$ or $F_{\text{cPAM}}$ is the larger force. This is achieved by the pretensioned spring. The forces in the edge case of $p_{\text{cPAM}} = p_{\text{max}}$ and $p_{\text{supply}} = 0$ are:
\begin{align}
    F_{\text{cPAM,max}} &= \frac{1}{2} 4 \pi \left( \frac{D_{\text{ball}}}{2} \right) ^2 p_{\text{max}} \\
    F_{\text{spring}} &= k  x_0 \geq F_{\text{cPAM,max}}
\end{align}
We use a spring with spring constant $k = 0.928\,\text{N/m}$ which results in a theoretical pretension distance $x_0 = 0.424\,\text{mm}$. If the valve comes close to the permanent magnet inside the tip mount, it must always open. This gives the following edge case with $F_{\text{supply,max}} = F_{\text{cPAM,max}}$:
\begin{align}
    F_{\text{magnet}} \geq F_{\text{spring}} + F_{\text{supply,max}}
    = \pi D_{\text{ball}}^2 p_{\text{max}}
\end{align}
The correct external magnet to achieve this was determined empirically. We use a neodymium N52 cylinder magnet (McMaster-Carr) of diameter 19.05\,mm and height 12.7\,mm.

The pretension on the spring is crucial for the valve to properly open and close in the desired working range. The pretension is adjustable to overcome mechanical imperfections in the spring and the 3D-printed parts and to allow for different working ranges. It is achieved by dividing the 3D-printed housing into multiple parts, which are threaded into each other. When tightening the thread, the length of the valve decreases and the pretension on the spring increases.

To manufacture the valve, the bottom, middle, and top parts are 3D-printed (Formlabs Form 3+ with Clear V4 material). The o-ring is sealed into the top part with a liquid sealant (Gear Aid Seam Grip). We then insert the spring and the magnetic ball into the middle part and glue the top part onto it. Subsequently, we tune the valve by threading the bottom part onto the middle part. Finally, the thread is locked using superglue (The Original Superglue Gel) and the valve is tested in a water bath. The complete process takes approximately five minutes of active work per valve, with a waiting time of three hours for the sealant to dry. One valve costs \$0.50 in a batch of 1000 valves.

\subsection{cPAM Design}

\begin{figure}[t]
    \centering
    \includegraphics[width=0.45
    \textwidth]{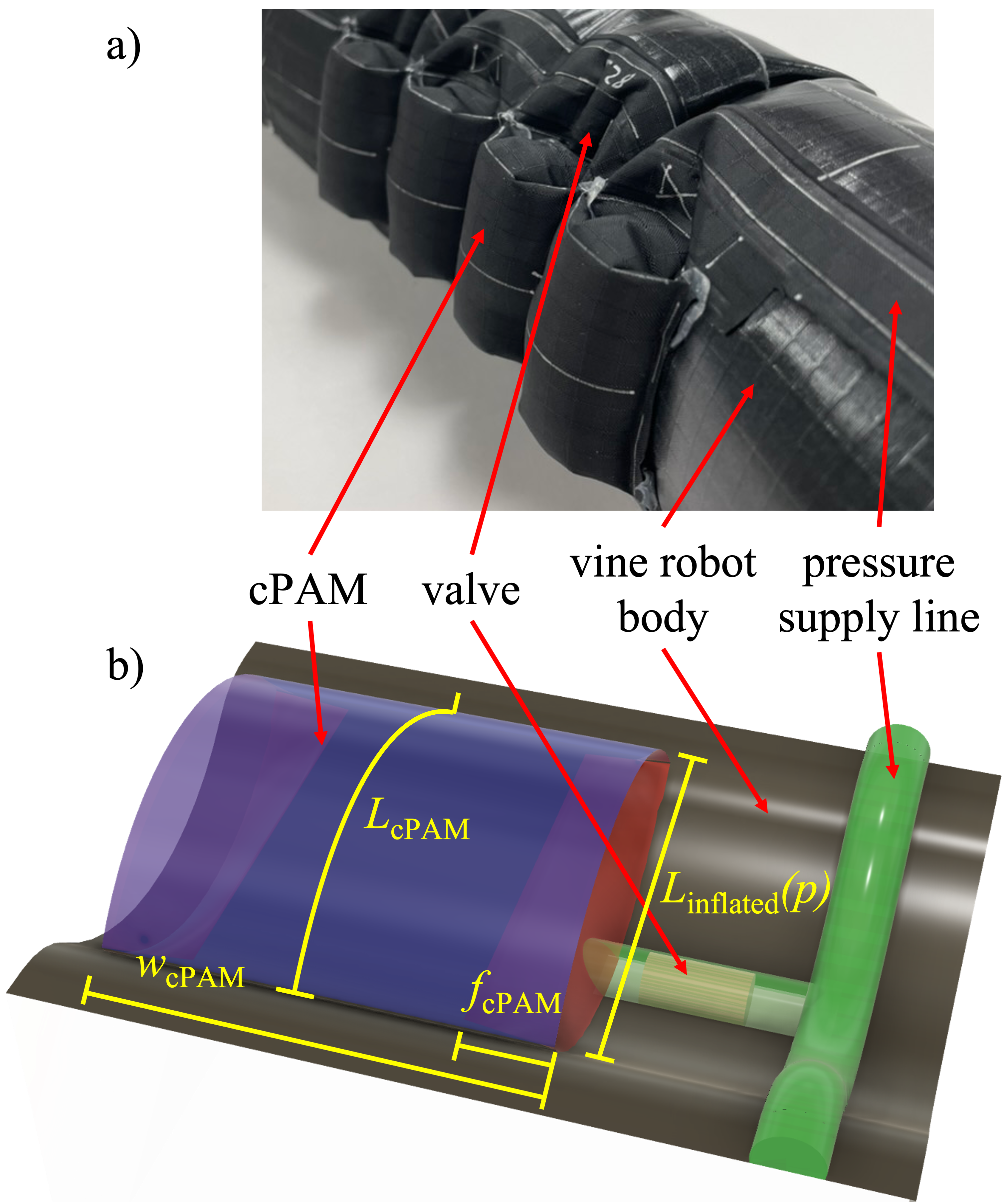}
    \caption{Details of the cPAM with the valve and the pressure supply line in a) the manufactured vine robot and b) a visualization. The integrated cPAM inflates to a cylinder due to the folded material on its sides. The pressure is supplied by the pressure supply line and controlled through the valve. The dimensions are $w_{\text{cPAM}} = 40\,\text{mm}$, $L_{\text{cPAM}} = 40\,\text{mm}$, and $f_{\text{cPAM}} = 10\,\text{mm}$.}
    \label{fig:integration}
\end{figure}

We propose an air pouch that inflates to a cylindrical shape, the cylindrical pneumatic artificial muscle (cPAM). It behaves like an ideal pouch motor \cite{niiyama2015pouch} because folded material on its sides prevents the edges from being constrained, similar to the folded pneumatic actuators in \cite{wang2022folded}. The cPAM is fully integrated into the vine robot. As in \cite{abrar2021highly}, the body of the vine robot is also a surface of the cPAM. \Cref{fig:integration} shows the cPAM integrated into the vine robot with the pressure supply line and the valve.

We assume the cPAM is flat in the deflated state and forms a cylinder in the inflated state. The fold of length $f_{\text{cPAM}}$ forms the ground area of the cylinder with diameter $D_{\text{cPAM}}$:
\begin{align}
    f_{\text{cPAM}} &= \frac{1}{2} D_{\text{cPAM}} = \frac{1}{2} \frac{2}{\pi} L_{\text{cPAM}}\\
    2 f_{\text{cPAM}} &\leq w_{\text{cPAM}}
\end{align}
We use $w_{\text{cPAM}} = 40\,\text{mm}$ and $L_{\text{cPAM}} = 40\,\text{mm}$, resulting in $f_{\text{cPAM}} = 12.74\,\text{mm}$. For simplicity, we use $f_{\text{cPAM}} = 10\,\text{mm}$.
The vine robot has a diameter of $D_{\text{vine}} = 80\,\text{mm}$.

\subsection{Vine Manufacturing}

The vine robot is made of 70 Denier Ripstop Nylon with a TPU coating on one side for air tightness (Quest Outfitters). The pressure supply line and the cPAMs are directly integrated into the vine robot by welding additional layers onto the vine robot body using an ultrasonic welder (Vetron 5064). 

The manufacturing process for the vine robot is as follows: First, we draw the outlines of the vine robot body and the cPAMs on the fabric and cut it to shape. Then, we mount, glue, and seal valves between two fabric layers. We weld the cPAM fabric along a prescribed pattern and check for air tightness by pressurizing each cPAM individually underwater to 40\,kPa. Finally, we weld the vine to form a tube. In the final prototype, pairs of cPAMs are connected to one valve to simplify manufacturing.

The supply system to control the vine robot is similar to the one shown in \cite{coad2019vine}. It includes a pressurized base with a spool for the uneverted vine robot material and two QB3 pressure regulators for the two pressure supply lines and one for the vine robot body.

\subsection{Motorized Tip Mount}

\begin{figure}[t]
    \centering
    \includegraphics[width=0.48
    \textwidth]{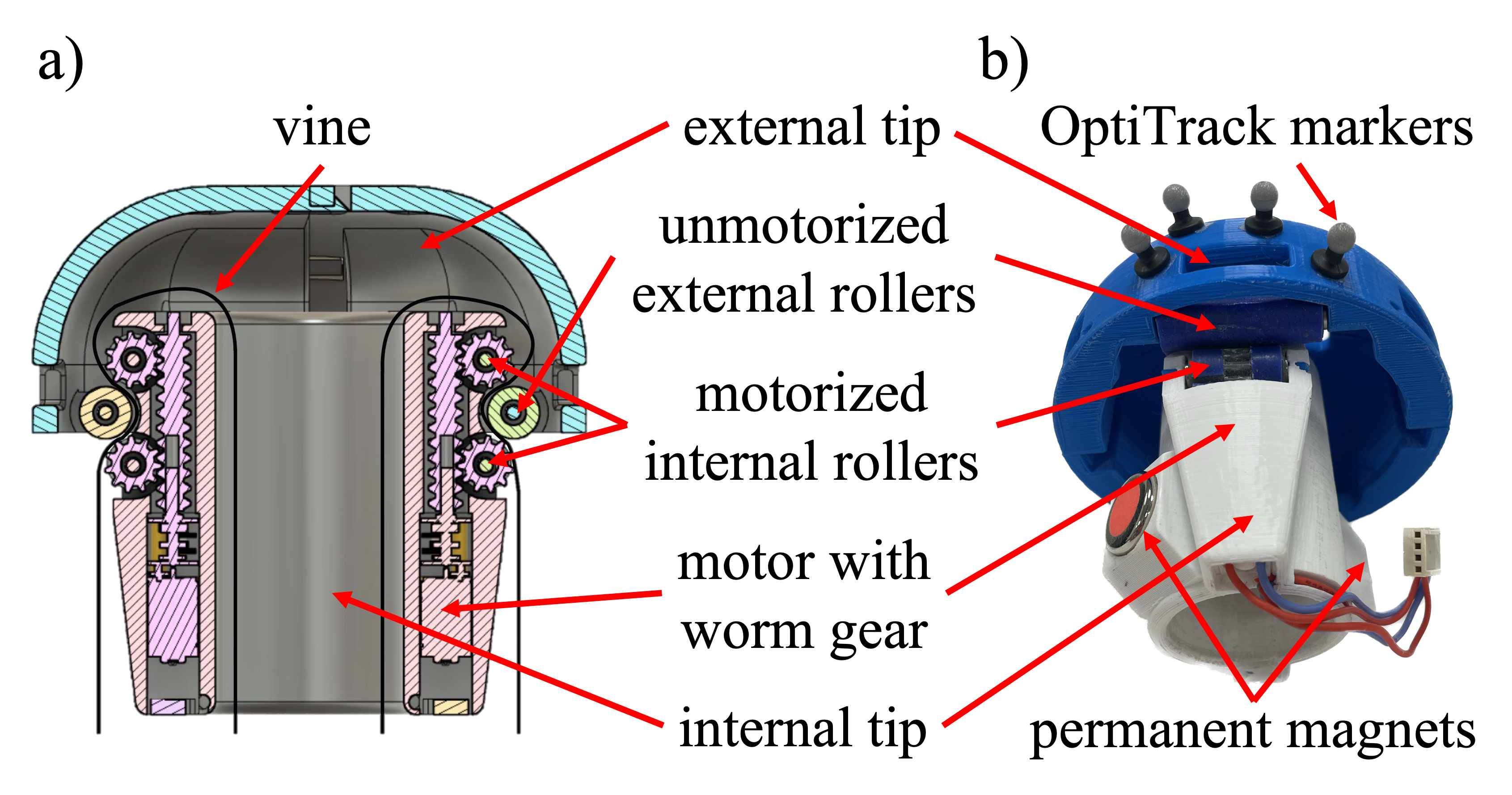}
    \caption{The motorized tip mount with actuated rollers. a) CAD intersection view from the side. b) The assembled tip mount from the back. The tip mount consists of an external and internal part and two sets of motorized rollers. It houses the permanent magnets to actuate the valves. OptiTrack markers enable tracking the vine robot's motion.}
    \label{fig:tip}
\end{figure}

We combine the idea of an interlocking tip mount using passive rollers \cite{der2021roboa} with the use of motorized rollers on the inner vine robot material \cite{coad2020retraction}. Our design has two sets of rollers on the outer vine robot material (\Cref{fig:tip}). Each set consists of two motorized and one passive roller. This interlocking mechanism allows the vine robot to evert evenly through the rollers and the tip mount. The motorized rollers are connected to the internal part and are driven by a high-geared motor (Pololu 1000:1 Micro Metal Gearmotor HPCB 6V). They control the continuous eversion and retraction speed and prevent buckling during retraction by relaxing the vine robot body material inside the tip mount. In contrast to previous retraction devices, the rollers act on the outer material of the vine robot. This prevents any interference of the rollers with the valves.
The tip mount carries the permanent magnets that actuate the valves. The magnets are inserted into the internal tip mount on the bottom left side (left actuator line) and the top right side (right actuator line).
Sensors such as cameras or microphones can be attached to the tip mount.

\section{Modeling}
\label{modeling}

\subsection{Maximal Bending}

The cPAM is designed to be a flat rectangle in the deflated state and a cylinder in the inflated state. The folded material on the sides forms the ground area of the cylinder. This is the working principle of an ideal pouch motor with a theoretical contraction of $\epsilon = 0.363$ \cite{niiyama2015pouch}. Under the assumption that the actuated side ($L_{\text{inflated}}(p)$) contracts and the unactuated side ($L_{\text{cPAM}}$) keeps its length, we can calculate the bending angle per length $q(p)/L_{\text{cPAM}}$ (\Cref{fig:pcc}):
\begin{align}
    \epsilon &= \frac{L_{\text{cPAM}} - L_{\text{inflated}}(p)}{L_{\text{cPAM}}} \\
    \frac{q(p)}{L_{\text{cPAM}}} &= \epsilon  \frac{360 \degree}{2 \pi D_{\text{vine}}}
\end{align}
With the theoretical contraction $\epsilon$ and $D_{\text{vine}} = 80\,\text{mm}$ we get a theoretical bending angle per length of $0.26\,\degree/\text{mm}$.
With a deflated cPAM length of $L_{\text{cPAM}} = 40\,\text{mm}$, we can achieve a theoretical bending of $10.4\,\degree$ per cPAM.

\subsection{Kinematic Modeling}

\begin{figure}[t]
    \centering
    \includegraphics[width=0.48
    \textwidth]{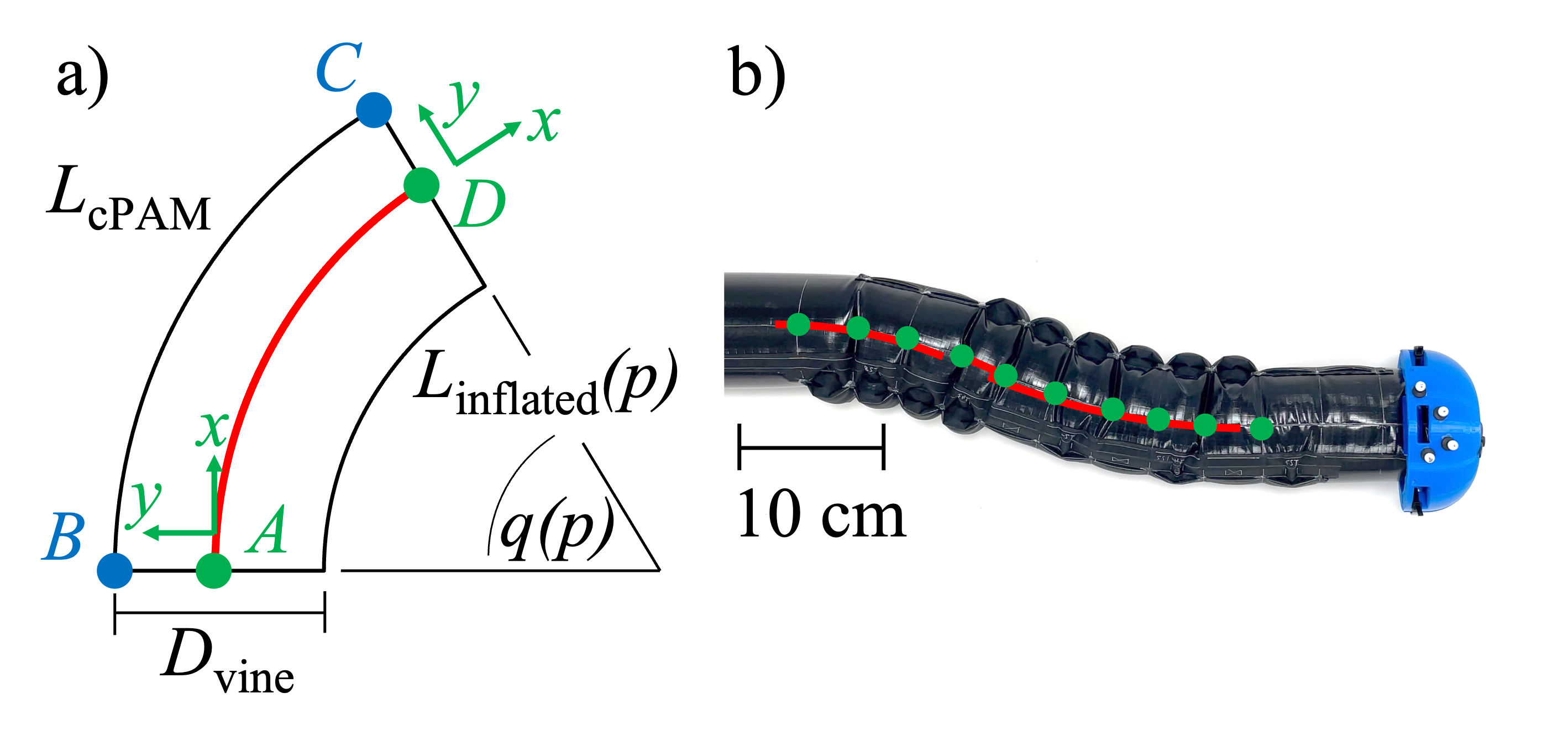}
    \caption{a) A vine robot segment over one cPAM is modeled with a constant curvature. The constant length is assumed on the unactuated side ($L_{\text{cPAM}}$). b) Piecewise constant curvature segments predict the vine robot path.}
    \label{fig:pcc}
\end{figure}

To predict the path of the vine robot, we assume that every cPAM creates a constant curvature. The vine robot is modeled with a piecewise constant curvature (PCC) approach \cite{della2018dynamic}. We experimentally create a lookup table to relate the pressure $p$ in a cPAM to the bending $q(p)$ (\Cref{fig:stability}). The key difference to previous PCC models is that for vine robots we assume the length of the unactuated side instead of the center line to be constant. This requires two additional transformations, from point A to B and from C to D (\Cref{fig:pcc}). The PCC transformation is performed between B and C. We can calculate the following transformation matrices depending on the bending $q(p)$. We use the corrected segment length $\Tilde{L}$ to account for the small length change in the manufactured vine robot caused by the insertion of the valve.
\begin{align}
    T_{\text{AB}} &= \begin{bmatrix} 1 & 0 & 0 \\ 0 & 1 & \frac{q}{\lvert q \rvert}\frac{D_{\text{vine}}}{2} \\ 0 & 0 & 1 \end{bmatrix},
    T_{\text{CD}} = \begin{bmatrix} 1 & 0 & 0 \\ 0 & 1 & \frac{-q}{\lvert q \rvert}\frac{D_{\text{vine}}}{2} \\ 0 & 0 & 1 \end{bmatrix}\\
    T_{\text{BC}} &= \begin{bmatrix} \cos{q} & -\sin{q} & \Tilde{L}\frac{\sin{q}}{q} \\ \sin{q} & \cos{q} & \Tilde{L}\frac{1-\cos{q}}{q} \\ 0 & 0 & 1 \end{bmatrix}\\
    T_{\text{AD}} &= T_{\text{AB}}  T_{\text{BC}} T_{\text{CD}}\\
    \Tilde{L} &= L_{\text{cPAM}} - (\frac{\pi}{2} - 1) D_{\text{valve}}
\end{align}

\section{Experimental Evaluation}
\label{sec:results}

\subsection{Bending and Shape Stability}

\begin{figure}[t]
    \centering
    \includegraphics[width=0.35
    \textwidth]{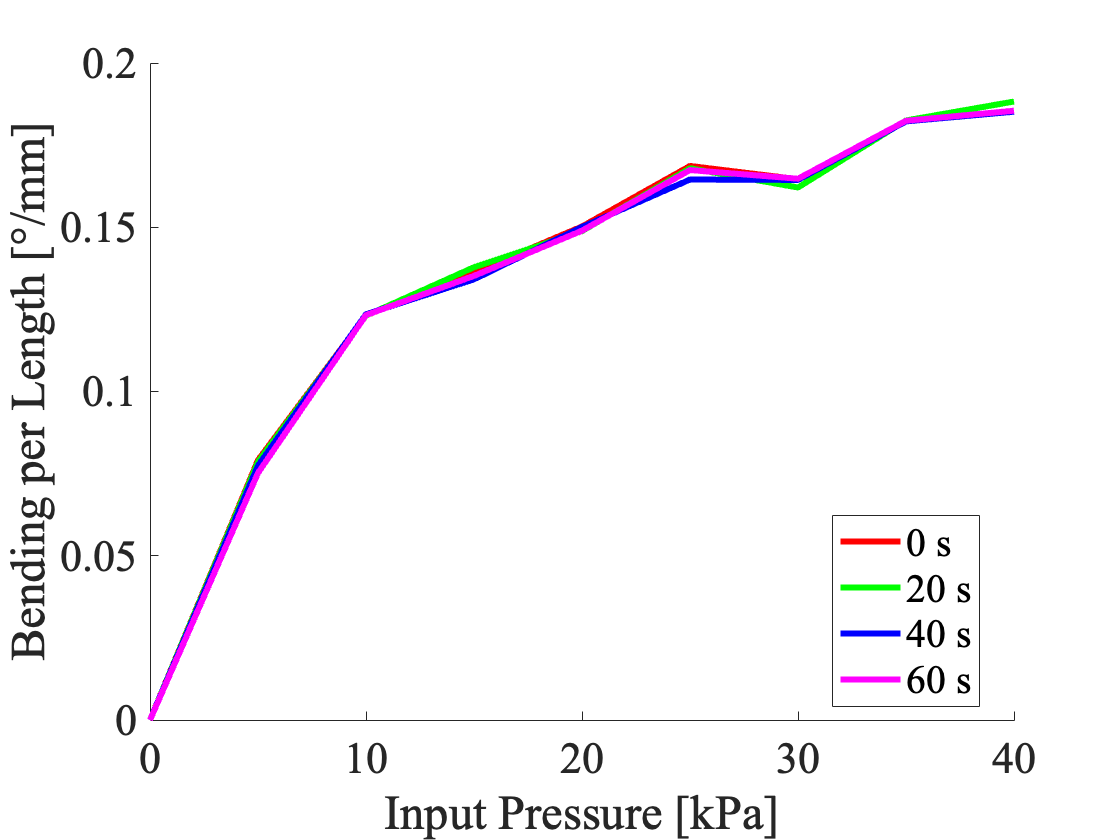}
    \caption{The bending angle per length in relation to the supplied pressure. The different curves show measurements at different times after setting the pressure in the pressure supply line to zero. The vine robot body pressure was constant at 3.7 kPa.}
    \label{fig:stability}
\end{figure}

We placed markers on the vine robot and investigated the bending characteristics using an Intel Realsense D415 camera. The results are shown in \Cref{fig:stability}. The vine robot was placed horizontally on the ground. We supplied a pressure to the pressure supply line and opened all valves (magnet positioned manually) to inflate all cPAMs. Then we set the pressure in the pressure supply line to zero and measured the bending angle per length over time. The bending increases with higher pressures and reaches its maximum at around 40\,kPa. The maximum bending is $0.185\,\degree/\text{mm}$. The vine robot stays stable in its position without a pressure in the pressure supply line. This means the system is airtight, and the valves close the cPAMs successfully.

\subsection{Growing into Shapes}
\label{automatic}

\begin{figure}[t]
    \centering
    \includegraphics[width=0.48
    \textwidth]{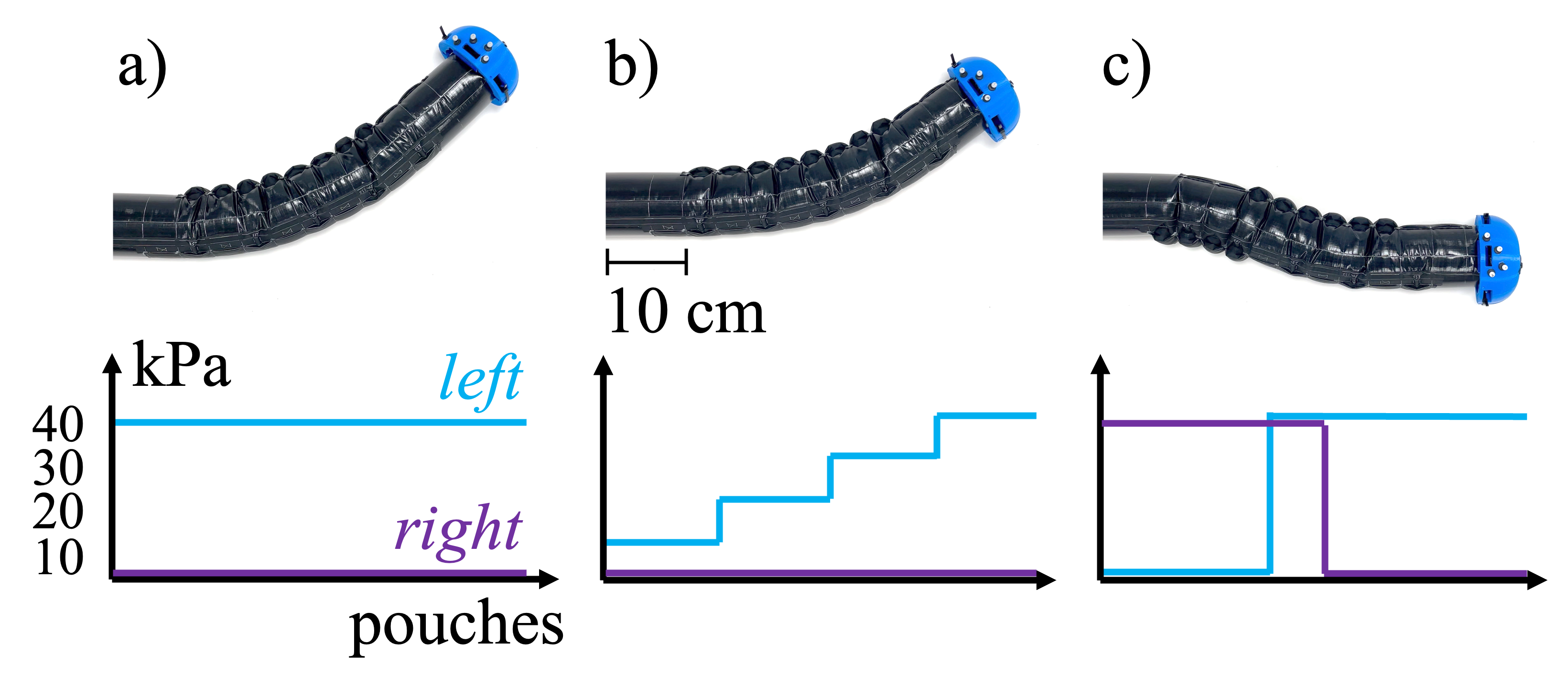}
    \caption{The vine robot grows into different shapes and holds them. The blue line indicates the pressure in the cPAMs on the left side, the purple on the right side. a) Left turn with constant pressure (40\,kPa) in all cPAMs. b) Left turn with increasing pressure (10, 20, 30, 40\,kPa). c) S turn with constant pressure (40\,kPa). First to the right over three cPAMs, then straight over one cPAM, then left over five cPAMs.}
    \label{fig:automated_picture}
\end{figure}

\begin{figure}[t]
    \centering
    \includegraphics[width=0.4
    \textwidth]{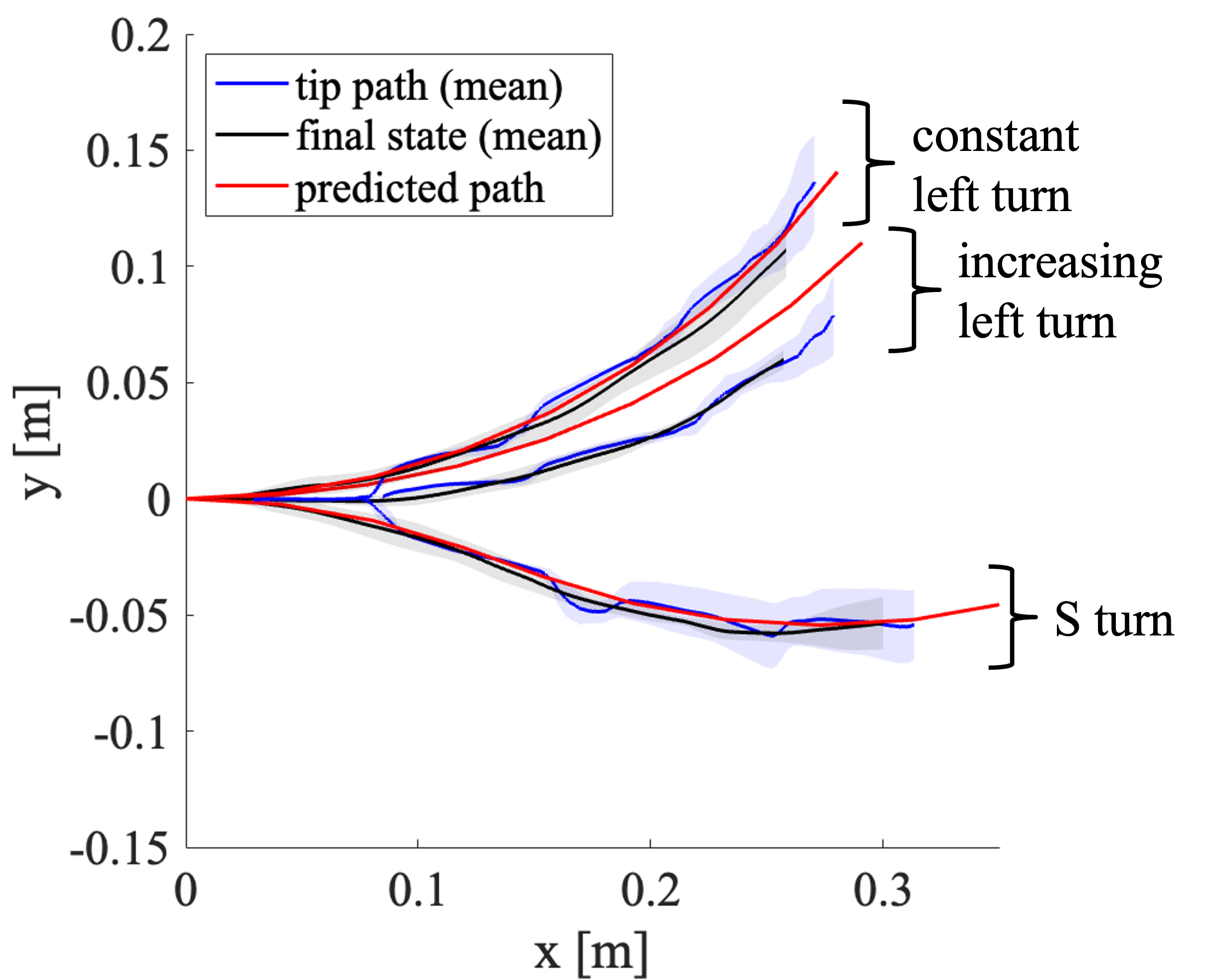}
    \caption{The vine robot growing into shape: The measured tip positions (blue), the final state of the vine robot (black), and the path predicted by the model (red). The lines represent the mean and the shaded areas show the standard deviation over five iterations.}
    \label{fig:automated_graph}
\end{figure}

\begin{figure}[t]
    \centering
    \includegraphics[width=0.325
    \textwidth]{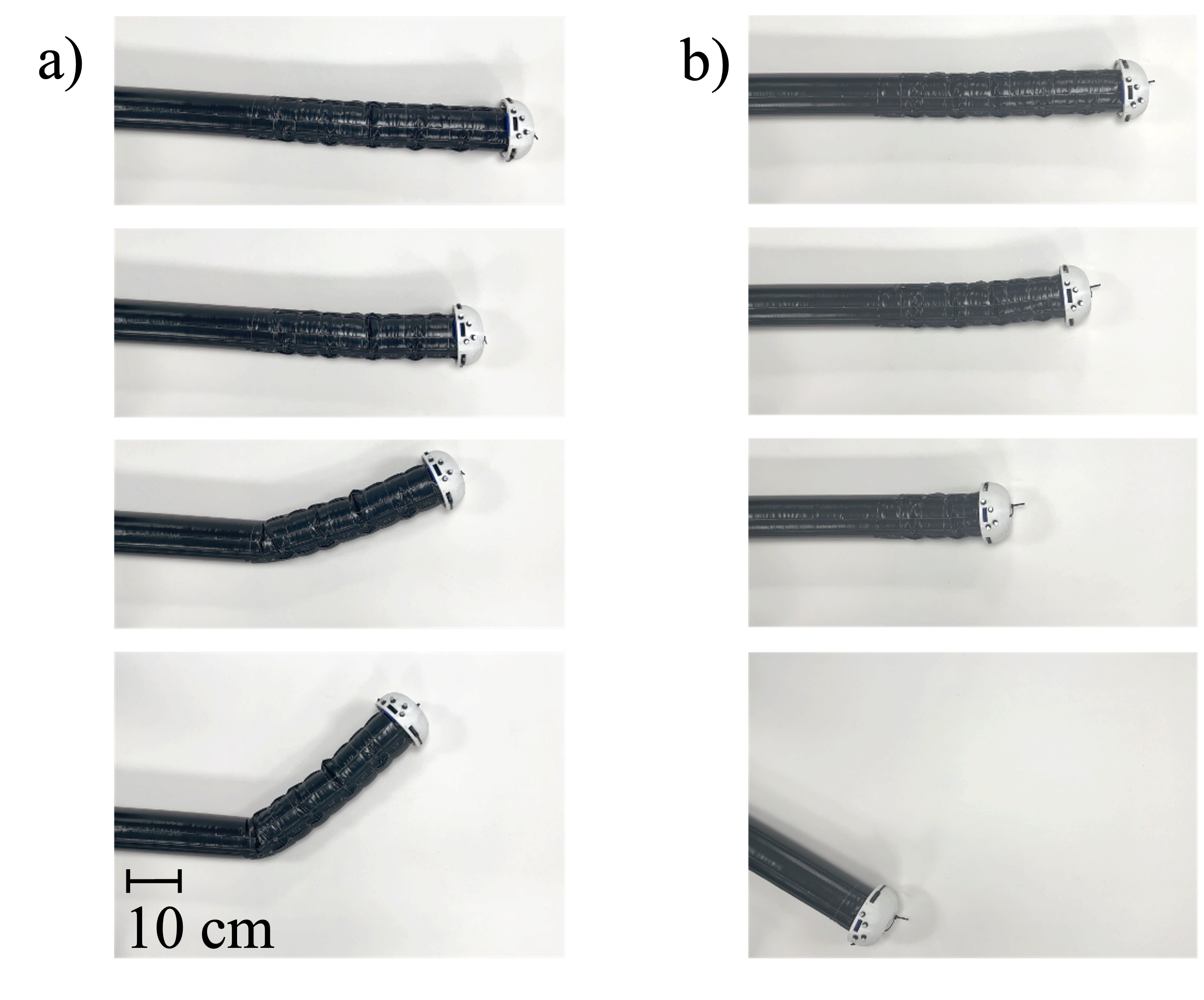}
    \caption{Example of retraction and resulting buckling with a) an unmotorized tip mount and b) the motorized tip mount. The unmotorized tip mount buckles earlier.}
    \label{fig:retraction}
\end{figure}

We performed a test where the vine robot grows into different shapes, without any manual mechanical  input. The valves were opened by the magnets in the tip mount and the motorized rollers in the tip mount controlled the eversion speed. The vine robot used in this experiment has eight cPAMs on each side, with two cPAMs per valve.
\Cref{fig:automated_picture} shows three pressure inputs and the resulting final state of the vine robot: A left turn with constant pressure (Figure 9a), a left turn with increasing pressure (Figure 9b), and an S turn (Figure 9c). \Cref{fig:automated_graph} shows the corresponding position measurements (OptiTrack) of the moving tip mount (blue), the final vine robot state (black), and the predicted path (red). Each test was repeated five times. We demonstrate that the vine robot can grow into distinct shapes depending on the steering direction and the supplied pressure. Each cPAM contributes a distinct bend and the supplied pressure changes the amount of bending (see the difference between the constant left turn and the increasing left turn). The direction and strength of a turn are thus controllable. 
The standard deviation resulting from five iterations is small compared to the vine robot's dimension. The standard deviation at the tip of the final vine robot state is $11.8$\,mm for the constant left turn, $4.1$\,mm for the increasing left turn, and $11.3$\,mm for the S turn. It shows that the vine robot can grow into the same shape reliably over several test iterations. The path of the tip (blue) matches the final state of the vine robot (black). We show that the vine keeps its position constant after growing into shape.
One interesting observation is that the final vine robot state (black) is smooth, but the tip mount path (blue) is jagged. This is because the tip undergoes an abrupt and discrete movement at the moment when a new cPAM is inflated.
The model (red) predicts the path within one standard deviation for the constant left turn and the S turn. The mean error distance between the model and the final vine robot state is $2.5$\,mm for the constant left turn and $3.1$\,mm for the S turn. The model overpredicts the bending of the increasing left turn with a mean error distance of $10.8$\,mm. This suggests that the experimentally determined pressure-to-bending relationship works well for high pressures but overpredicts bending for lower pressures.
More experiments could better characterize the pressure-to-bending relationship. Also, the model predicts the path to be slightly too long. In reality, cPAMs are often a little shorter due to manufacturing imprecision.

\subsection{Motorized Tip Mount \& Retraction}
\begin{figure}[t]
    \centering
    \includegraphics[width=0.325
    \textwidth]{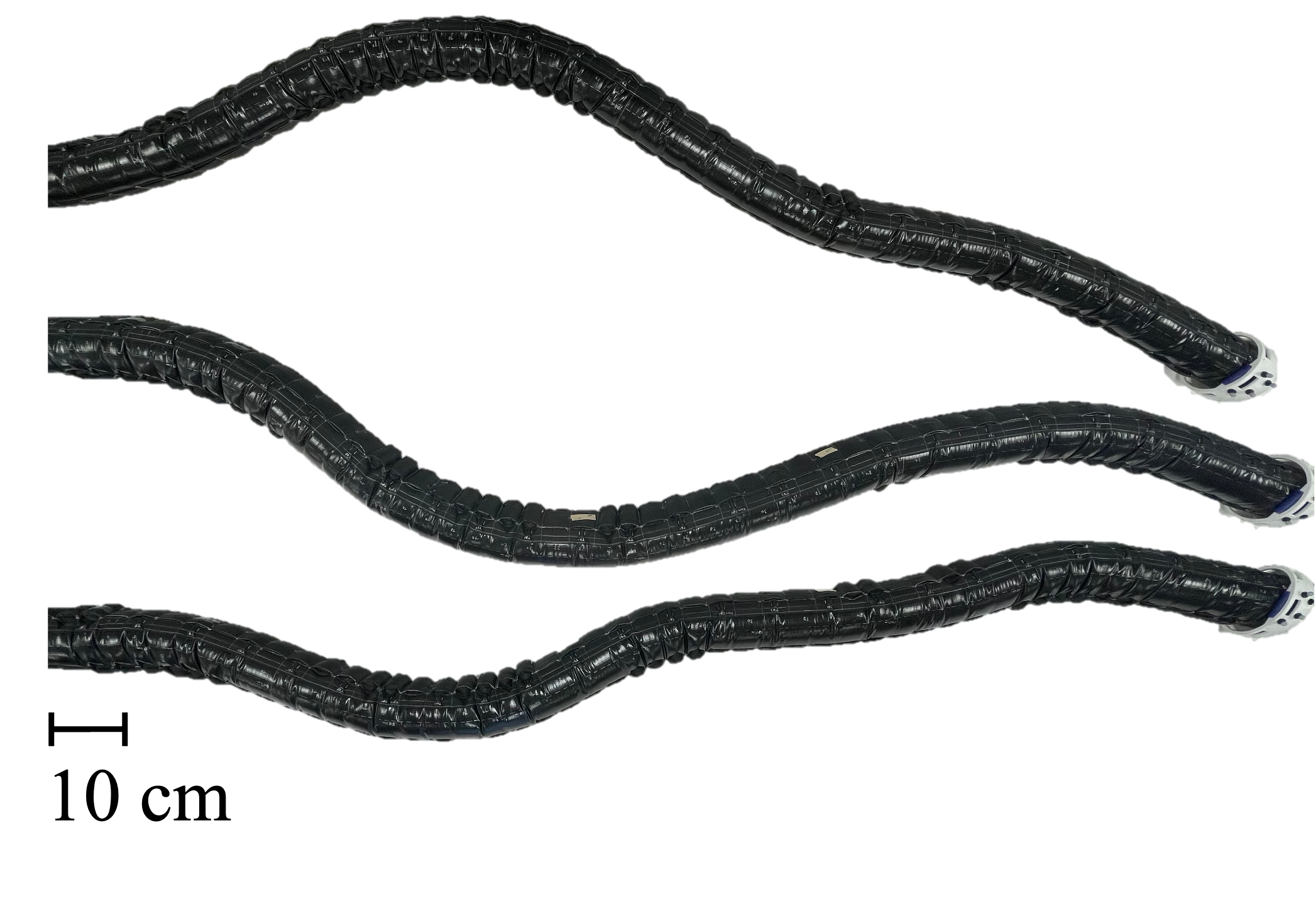}
    \caption{A 2\,m long multi-segment vine robot in three different shapes. In this test, the cPAMs were inflated manually.}
    \label{fig:longvine}
\end{figure}

The tip mount with motorized rollers achieves a constant eversion speed. An unmotorized version of our tip mount, where the rollers are not connected to the motor, moves faster but with abrupt forward jumps. This makes actuating the valves difficult because the magnets in the tip mount skip valves during jumps. 
The pressure required for eversion is lower for the motorized tip mount (3\,kPa versus 6\,kPa for the unmotorized version). This is particularly important because we can achieve higher bending angles with a lower vine robot body pressure.

For retraction, we experienced that buckling occurs less often with the motorized tip mount than with the unmotorized tip mount (\Cref{fig:retraction}). This shows that retraction works with motorized rollers on the vine robot's outer material rather than on the inner material. The rollers push the outer material into the front of the tip mount, and the material can be pulled back with a smaller force. However, our design still showed some instances of buckling. This is likely due to the manufacturing. The 3D-printed tip mount is not stiff enough, limiting the force the rollers can apply. Building a more robust version with stiffer parts and metal worm gears will improve the retraction performance significantly. Also, the synchronization between the rollers and the pull-back force from the spool in the base was not ideal. Addressing these two issues will make retraction more reliable.

\subsection{Long Vine Demonstration}

We manufactured a 2\,m long vine robot with the same cPAM and body dimensions as in \Cref{automatic}. The manufacturing process was not precise enough for this vine robot to simultaneously grow and inflate the cPAMs. The vine robot was slightly twisted along its center axis, which made precise alignment with the tip mount difficult over long distances. Also, a few of the 50 cPAMs leaked, and some valves did not shut properly, which led to unwanted cPAM inflation. Although this vine robot reveals the current limitations in scaling up manufacturing, it shows how its shape can be chosen with great flexibility~(\Cref{fig:longvine}). In this test, the vine robot was first fully everted without inflating the cPAMs. Selected cPAMs were then inflated by manually placing the magnet on the corresponding valve. After bringing the vine robot into its shape, we set the pressure in the pressure supply line to zero and the vine robot held its shape constant with no visible changes after waiting for 15 minutes.

\section{Conclusion \& Future Work}

We present a novel multi-segment vine robot that can grow into various shapes without contacting the environment. The principle allows for high degrees of freedom with only two pressure control inputs. A manufactured prototype was able to repeatably grow along different paths. It is airtight and can hold its shape without an active air supply to the pouches. 

A permanent magnet inside the tip mount selectively actuates single pouches or a connected set of pouches. The user only has to input the pressure to the robot during operation and does not need to control an additional shape-locking or stiffening system. The design is reversible because the pouches can deflate during retraction. 
It can be adapted to different requirements by varying the pouch type, the pouch size, the material, or the number of pouches connected to one valve. The implementation is cheap because it only requires valves and magnets in addition to the vine robot fabric.

The developed 3D-printed magnetic valve is actuated by an external magnet. The valve is cheap, easy to manufacture, and can be tuned for different pressure ranges.

With the cylindrical pneumatic artificial muscle (cPAM), we introduce a new vine robot air pouch that is integrated into the vine robot body. It achieves high bending angles because folded material on the sides of the cPAM allows it to form a cylindrical shape. The cPAM can also be used with non-multi-segment vine robots to improve their performance.

The developed tip mount uses motorized rollers acting on the outer material of the vine robot. It controls the eversion speed, which is essential for a reliable valve actuation, and decreases the likelihood of buckling during retraction. It allows for having stiff parts inside the vine robot and decreases the necessary eversion pressure.

Finally, we present a piecewise constant curvature model adapted for vine robots that accurately predicts a multi-segment path from the measured pressure-to-bending relationship of the cPAM.

Possible applications include exploring confined spaces without relying on environment contact to steer (archeology or search and rescue), collecting environmental samples along predefined paths, and creating three-dimensional structures in low-gravity environments (space).

The biggest current limitation is the precision in manufacturing. The manufacturing methods for this work were heavily dependent on human skills and 3D-printing resolution. For the principle to work properly, the valves, the welding of the fabric layers, and the integration of the valves into the fabric have to be fully airtight. Furthermore, the manual manufacturing led to a misalignment between the magnets in the tip mount and the valves for the long vine robot.
Future work will therefore focus on more precise manufacturing methods to increase robustness and reliability to then realize longer multi-segment vine robots. Such methods could include machine-guided welding techniques and injection-molded valves with a weldable exterior.

We would like to extend the design to a vine robot that can steer in the three-dimensional space by adding a third strand of cPAMs, a third pressure supply line, and a third permanent magnet.
To enable an automated deployment, we will develop a feedback controller.

Finally, the idea of selectively actuating different pouches is not only limited to vine robots but can be extended to other soft continuum robots.








\section*{ACKNOWLEDGMENT}
The authors like to thank Nathaniel Agharese, Brian Do, and Rianna Jitosho for design discussions, Zhenishbek Zhakypov for discussions on valve designs, Nicolas Aymon and Betim Djambazi for discussions on pouch designs and manufacturing, and Sarina Mayya for her contribution to build the tip mount.


\bibliographystyle{IEEEtran}
\bibliography{IEEEabrv,references}

\begin{acronym}
\acro{imu}[IMU]{Inertial Measurement Unit}
\end{acronym}

\end{document}